\documentclass[letterpaper, 10 pt, conference]{ieeeconf}  

\usepackage[top=54pt,bottom=54pt,left=54pt,right=54pt]{geometry}

\usepackage{amssymb, amsmath, amsthm}
\usepackage{siunitx}

\usepackage{cite}
\usepackage{graphicx}
\usepackage{multirow}
\usepackage{makecell}

\usepackage{blindtext}
\usepackage{algorithm}
\usepackage{algpseudocode}
\usepackage{hyperref}
\usepackage{verbatim}

\usepackage{makecell}
\usepackage{tikz}
\usetikzlibrary{fit}

\IEEEoverridecommandlockouts                              

\title{\LARGE \bf
Accurate Simulation and Parameter Identification of Deformable Linear Objects using Discrete Elastic Rods in Generalized Coordinates
}

\author{Qi Jing Chen$^{1}$, Timothy Bretl$^{2}$, and Quang-Cuong Pham$^{3}$
\thanks{*Github repo: \url{https://github.com/qj25/adapteddlo_muj} (Videos: \url{https://youtu.be/QY0B5IChghw})}
\thanks{$^{1}$Nanyang Technological University, School of Mechanical and Aerospace Engineering, $^{2}$University of Illinois Urbana-Champaign,
$^{3}$Eureka Robotics, Singapore
	}%
}

\newcommand{\cmmnt}[1]{\ignorespaces}
\begin{document}

\maketitle

\begin{abstract}
	
This paper presents a fast and accurate model of a deformable linear object (DLO) -- e.g., a rope, wire, or cable -- integrated into an established robot physics simulator, MuJoCo. Most accurate DLO models with low computational times exist in standalone numerical simulators, which are unable or require tedious work to handle external objects. Based on an existing state-of-the-art DLO model -- Discrete Elastic Rods (DER) -- our implementation provides an improvement in accuracy over MuJoCo's own native cable model. To minimize computational load, our model utilizes force-lever analysis to adapt the Cartesian stiffness forces of the DER into its generalized coordinates. As a key contribution, we introduce a novel parameter identification pipeline designed for both simplicity and accuracy, which we utilize to determine the bending and twisting stiffness of three distinct DLOs. We then evaluate the performance of each model by simulating the DLOs and comparing them to their real-world counterparts and against theoretically proven validation tests.

\end{abstract}


\section{INTRODUCTION}
Deformable linear objects (DLOs) appear in a wide range of domains, including polymer physics \cite{wiggins2005exact}, musculoskeletal modeling \cite{zhang2019modeling}, hair simulation \cite{bertails2006super}, and DNA mechanics \cite{sereda2010evaluation}. Industrial applications include suturing in the medical field \cite{joglekar2024autonomous} and wire harness assembly in manufacturing \cite{zhou2020practical}. The robotic manipulation of DLOs in these contexts is an active area of research that demands precise control strategies. Key challenges include the complex, nonlinear dynamics of DLOs, their sensitivity to both internal and external parameters, and their systems' underactuated nature \cite{zhu2021challenges}.

In recent years, machine learning has gained popularity as a solution to such problems. Although they have the potential to overcome the shortcomings of classical control techniques, their implementation presents inherent difficulties. Training requires a large amount of data, especially so for complex tasks, and data collection in real setups could prove to be tedious and sometimes infeasible. The use of simulations to gather a diverse set of data is, therefore, a crucial component of the learning approach. To ensure that policies learned from simulated data perform well and are generalizable, simulations which are able to easily and accurately replicate real DLO dynamics are vital.

Our paper presents an accurate DLO model in MuJoCo~\cite{todorov2012mujoco}, an established physics simulator. Our model is an adaptation of the Discrete Elastic Rods (DER)~\cite{bergou2008discrete} theory into its generalized joint coordinates. Using results on computational speed, we show that this approach is more suitable for integration with the MuJoCo simulator as compared to direct implementation of the theory. We assume inextensibility of the DLO, which we believe to be a fair assumption for practical applications in wire manipulation and medical suturing. Through proven validation tests, our model is assessed and compared with the existing native cable model in MuJoCo which utilizes a stiffness model where torque responses are linearly related to bending and twisting deformations. We designed a novel parameter identification pipeline that emphasizes both simplicity and accuracy, using two separate tests to independently determine the bending and twisting stiffness of a DLO. We evaluate the consistency of the proposed method, thereby assessing its suitability for evaluating three distinct DLOs. We use the identified stiffness parameters to evaluate both models across a diverse set of four representative poses for three distinct DLOs, where one end of the DLO is fixed and the other is manipulated by a robot arm. With minimal decrease in computational speed, our adapted DLO model generally outperforms the native model.

\begin{figure}[t]
	\centering
	\includegraphics[width=0.45\textwidth]{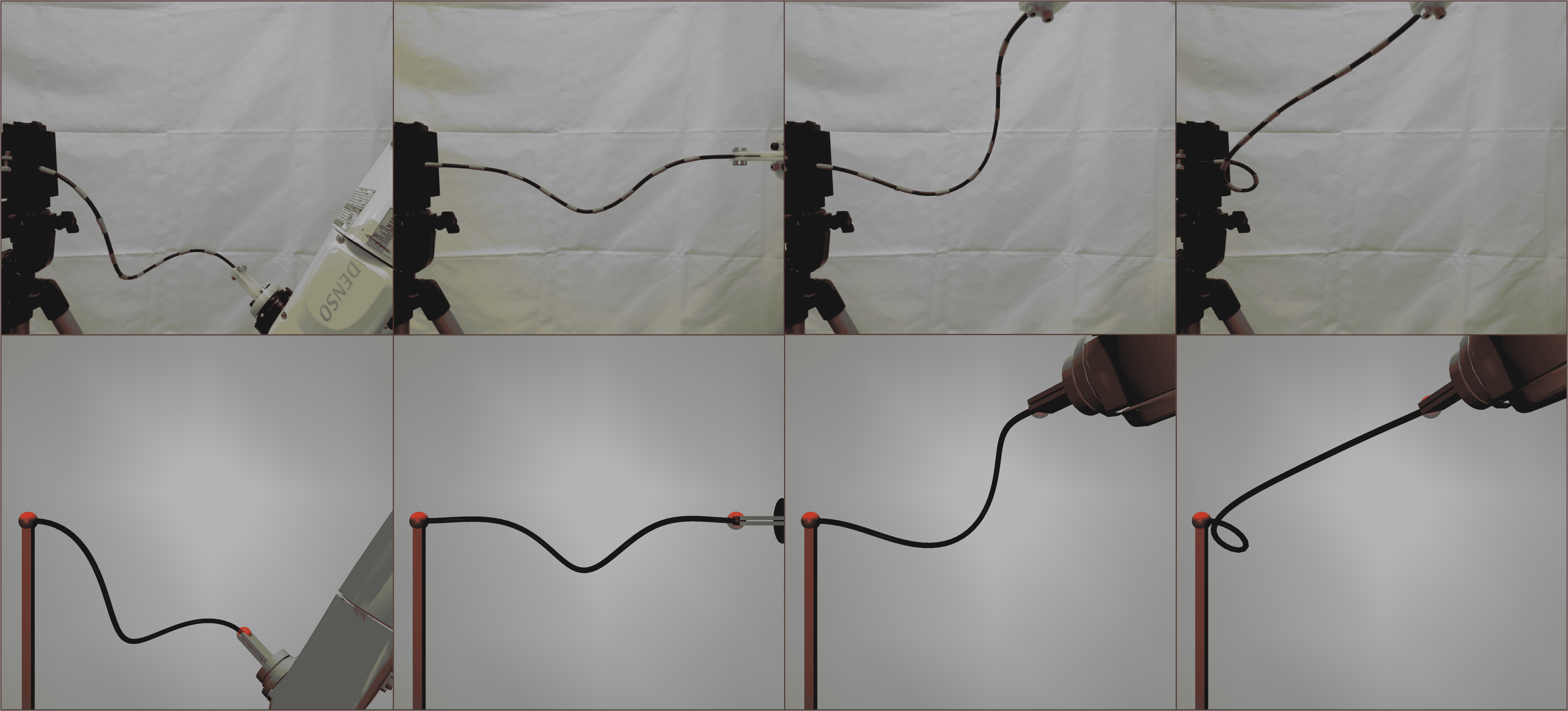}
	\caption{Comparisons of our adapted DER model (bottom) with a real black DLO (top) in 4 different poses. The left end of the DLO is held fixed (and twisted axially by a fixed amount) while the other end is manipulated by a robot arm. Stiffness parameters of the simulated models are determined using our novel parameter identification pipeline.}
	\label{fig:realexpsnapshots}
	\vspace{-0.3cm}
\end{figure}
\subsection*{Contribution and organization of the paper}

The contribution of our paper is threefold. First, we formally define the stiffness moduli used in the DER formulation with respect to more widely recognized physical and material parameters, a detail which was missing from the original paper \cite{bergou2008discrete} (Section~\ref{section:stiffmod}).

Second, we present a generalized coordinate representation of the Cartesian stiffness forces from the DER theory, which is better suited for integration into MuJoCo and reduces computational load. In addition, the adoption of DER theory's quasistatic treatment of the rod's material frame twist improves stability and minimizes unnatural oscillations of the DLO simulation. Following that, we conduct two validation tests presented in the original literature\footnote{Video of simulation validation for adapted model: \url{https://youtu.be/QY0B5IChghw}} \cite{bergou2008discrete} on our model and MuJoCo's native cable model. Details can be found in Section~\ref{section:adaptmod} and Section~\ref{section:perf}.

Third, we introduce a novel parameter identification pipeline which emphasizes simplicity and accuracy to determine bending and twisting moduli of a DLO. The pipeline comprises two test: one to determine the bending stiffness, the other to assess the twisting-to-bending stiffness ratio. Using the identified parameters, we evaluate the two simulation models (ours and the native model) against real experiments where one end of a DLO is fixed and the other end manipulated by a robot arm into 4 different poses. For more details, see Section~\ref{section:realexp}.


\section{RELATED WORK}
This section reviews the literature on DLO manipulation, simulation, and parameter identification.

\subsection{Manipulation}
Manipulation of DLOs is difficult because of their complex dynamics and underactuated nature. Numerical simulations have been used to approximate DLO configuration \cite{lamiraux2001planning,moll2006path} to more accurately determine their geometric state. By proving that the set of all local solutions of a Kirchhoff elastic rod belongs to a smooth manifold of finite dimension \cite{bretl2014quasi}, sampling-based algorithms were used to traverse that path-connected space \cite{borum2015free} to achieve different rod shapes. These method are restricted by assumptions, including known stiffness parameters and the absence of external forces. Learning from demonstrations \cite{nair2017combining} and from images \cite{wang2019learning,zhang2021robots} can overcome this but creates a black-box policy making it difficult to understand the conditions under which failure occurs. Through learning the delta dynamics of the system from simulation, \cite{chi2022iterative} successfully carries out a dynamic rope flinging task capable of adapting to unseen rope types. Our work aims to reduce the need for real experiments by estimating DLO stiffness parameters through a novel pipeline and providing an accurate simulation which can be easily integrated with robot manipulators.

\subsection{Simulation}
DLOs have many different representations in simulation. Mass-spring systems are a common way to model DLOs~\cite{lloyd2007identification}. Despite their simplicity, an interpretation of the system parameters of these models as physical properties is sometimes lacking. To achieve the desired dynamic behavior could require substantial effort in tuning. Models like MuJoCo's native cable~\cite{todorov2012mujoco} have a linear torque response to deformations, which are not consistent with Kirchhoff rod theory. This limits its ability to accurately model DLOs, especially at large deformations. Position-based models \cite{umetani2014position} are popular in animation for visual plausibility, but lack physical accuracy. The Discrete Elastic Rod (DER) model \cite{bergou2008discrete} uses a concept of minimum energy to arrive at a stable configuration of the DLO, while having an explicit representation of the centerline which eases collision handling, and a quasistatic treatment of the material frame which can improve stability and performance of the simulation. In addition to bending and twisting, the Cosserat rod theory \cite{cosserat1909theorie,spillmann2007corde} include stretching and shearing to model elastic rods which have a wider array of applications. Our paper uses the DER theory to provide an accurate DLO model, where stretching and shearing are assumed to be negligible. More details will be discussed in Section \ref{section:ws}.

\subsection{Parameter identification}
Static parameter identification methods for DLOs~\cite{bartholdt2021parameter,liu2023robotic} compare equilibrium data from real and simulated experiments to determine physical parameters. Many of these approaches simplify the modeling of DLOs by omitting twist, which limits their ability to accurately estimate twist stiffness—an essential property for predicting DLO movement. Optimal control techniques have also been adopted to solve for a set of parameters from a real trajectory~\cite{caldwell2014optimal,yang2022learning}. The integration of machine learning in parameter identification has enabled for concurrent update of the DLO physical parameters during shape control~\cite{caporali2024deformable}. These methods are centered around manipulation and do not attempt to carry out parameter identification directly. Our work introduces a novel parameter identification pipeline which accounts for and accurately predicts both bending and twist stiffness. This pipeline is straightforward to implement, featuring a simple setup which only requires a set of 3-D printed apparatus and a depth camera.


\section{SIMULATION}
\label{section:ws}

In this paper, the DLO model will be simulated with the Discrete Elastic Rods (DER) \cite{bergou2008discrete} theory. This section will briefly introduce the DER model, details about the stiffness moduli used, and how the theory is being adapted into generalized coordinates for better integration with MuJoCo.

\subsection{Discrete Elastic Rods}
The DER theory splits a DLO into discrete sections with which its dynamics can be analyzed. The explicit representation of the centerline in the model allows for easy communication of simulation data to the model, without additional computation. Nodes are located at the joining point between adjacent sections. At each time step, the model calculates the Cartesian force on each node of the discretized DLO as the negative of the derivative of the energy with respect to the node position. Generally, the force acts in a direction which would result in the greatest decrease in the overall potential energy. The formulas are as follows:
\begin{equation}
	\label{eq:fi0}
	\ \vec{F_{i}} = -\frac{dE(\Gamma)}{dx_{i}},
\end{equation}
\begin{equation}
	\label{eq:fi1}
	\ \frac{dE(\Gamma)}{dx_{i}} = \frac{\partial E(\Gamma)}{\partial x_{i}} + \sum_{j=0}^{n} \frac{\partial E(\Gamma)}{\partial \theta^{j}} \frac{\partial \theta^{j}}{\partial x_{i}},
\end{equation}
where $F_{i}$ represents the force on node $i$, $E(\Gamma)$ is the energy for adapted framed curve $\Gamma$, $\theta^{i}$ is the angle required to rotate the Bishop frame into the material frame at section $i$, and $n$ is the number of discrete sections. The bending ($\frac{\partial E(\Gamma)}{\partial x_{i}}$) and twisting ($\sum_{j=0}^{n} \frac{\partial E(\Gamma)}{\partial \theta^{j}} \frac{\partial \theta^{j}}{\partial x_{i}}$) components in the energy differential are directly proportional to the stiffness moduli $\alpha$ and $\beta$, respectively. More details can be found in the original paper~\cite{bergou2008discrete}.

\subsection{Stiffness moduli}
\label{section:stiffmod}
In the DER model, the bending and twisting stiffness moduli are represented by $\alpha$ and $\beta$, respectively. Although these moduli appear in the governing equations, their connection to more widely recognized physical and material parameters is not explicitly discussed. We establish that connection in this section by expressing the moduli as $\alpha = EI$ and $\beta = GJ_T$, where $E$ is Young's modulus and $G$ is the shear modulus. Here, $I$ denotes the second moment of area of the DLO cross-section (assumed constant) about the axis of interest, and $J_T$ is the torsion constant for the section.

\subsection{Implementation}

\paragraph{Modeling the DLO}
Within the simulation, the DLO is modeled as a continuous chain of discrete capsules (one of the base object types that can be simulated in MuJoCo) which are joined at the nodes by ball joints, allowing for rotational but not translational motions. Damping forces in the joints are computed implicitly at each simulation step based on a user-specified damping coefficient. Stiffness torques computed from our adapted model (detailed in Section~\ref{section:adaptmod}) are applied at the joints. Stability and performance of the simulation can be adjusted with the damping coefficient, stiffness values, and time step size.

\paragraph{Constraints-handling}
To ensure the inequality constraint is respected, our work will make use of the constraint solver within MuJoCo, where a convex optimization problem is solved at each simulation step and the global solution is the DLO displacement at the next step.

\paragraph{Integration of the program}
For this paper, the model is integrated and tested in a MuJoCo simulation environment created with its native python bindings. At each time step, the \verb!update_torque! function is called which updates the new twist angle, $\theta_n$, and node positions, $x_i$, into the program. From these updated values, the centerline forces on the DER are calculated. These forces are converted to their equivalent torques, then applied to the DLO's ball joints using the MuJoCo variable, \verb!qfrc_passive!. Open-source code of the DLO model in MuJoCo is available on GitHub: \url{https://github.com/qj25/adapteddlo_muj}. For users working with MuJoCo’s original C++ API, an additional plugin is provided that integrates directly with MuJoCo’s source code.

\subsection{Adapted model in MuJoCo}
\label{section:adaptmod}
The DER theory simulates stiffness through Cartesian centerline forces. Direct application of these forces into MuJoCo results in querying for $n+1$ distinct Jacobians (one for each node) at each simulation step, which is a computationally expensive process. Our method converts the Cartesian forces into joint torques through moment-based force-lever analysis. Two reasons make this conversion possible. One, the net stiffness forces acting on the DER system is zero. Two, the discrete sections are connected using ball joints. The conversion of Cartesian forces into joint torques can be achieved by calculating the moment of each force about a corresponding joint, $\vec{\tau}_{i,j} = \vec{F}_{j} \times \vec{X}_{i,j}$ where $\vec{\tau}_{i,j}$ is the torque contribution at joint $i$ due to the stiffness force $\vec{F}_{j}$ applied at joint $j$. $\vec{X}_{i,j}$ is the vector distance from joint $i$ to $j$. In MuJoCo, the kinematic tree is built from a single parent body into subsequent child bodies. Joint torques input into the simulation application programming interface (API) can only work from the parent to child body. The swap of torque application body can be done with a simple negative sign (e.g., torque by A+1 on A is equivalent to the negative of torque by A on A+1). Since the kinematic tree is built from piece $0$ to $n$, torques in the joints can only be applied from piece $i$ to $i+1$. Therefore, $\vec{T}_{i,j}$, the joint torque contribution by $\vec{F}_{j}$ on joint $i$ in the simulation, is as follows,
\begin{equation}
	\label{eq:f2t_api}
	\vec{T}_{i,j} = \vec{F}_{j} \times \vec{D}_{i,j}
\end{equation}
such that 
\[
\vec{D}_{x,y} = 
\begin{cases}
	\vec{X}_{x,y},& \text{if } x < y\\
	\vec{X}_{y,x},              & \text{otherwise}.
\end{cases}
\]
and the overall torque on joint $i$ is,
\begin{equation}
	\label{eq:f2t_sum}
	\ \vec{T}_{i} = \sum_{j=0}^{n+1} (\vec{F}_{j} \times \vec{D}_{i,j}).
\end{equation}

Our adapted model provides a significant decrease in computational time as compared to direct application of the DER model as shown in Section~\ref{section:simuspd}.

\section{PERFORMANCE}
\label{section:perf}
Two important performance metrics for the simulation are computational speed and accuracy. In this section, we will address both aspects, then compare each model's performance against real experiments in Section~\ref{section:realexp}. All tests are carried out on an Intel i5-12400 6-cores CPU, with 32GB RAM.

\subsection{Simulation speeds}
\label{section:simuspd}
\begin{figure}[t]
	\centering
	\includegraphics[width=0.40\textwidth]{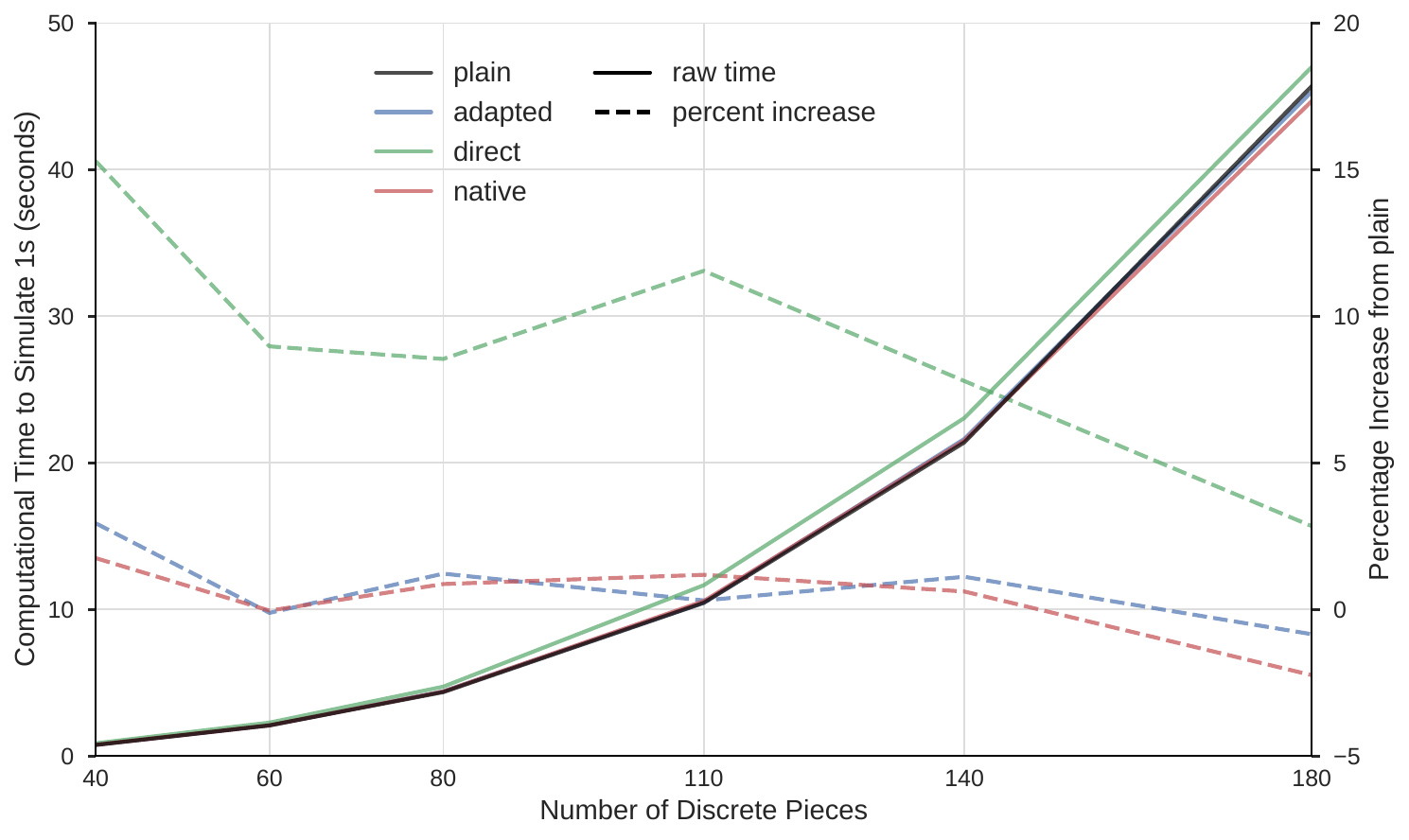}
	\vspace{-0.2cm}
	\caption{Computational time for each model across different $n$}
	\label{fig:computetime}
	\vspace{-0.4cm}
\end{figure}

The computational speeds for three different DLO models -- our adapted model (\verb!adapted!), the native MuJoCo cable model (\verb!native!), and the model which applies the Cartesian stiffness forces of the DER theory directly (\verb!direct!), will be compared against the kinematic chain model simulated without additional stiffness calculations (\verb!plain!). The computational times are shown in Fig.~\ref{fig:computetime} along with their percentage increase compared to \verb!plain!. The speed tests are a measure of the real computation time taken (average of 10000 time steps) to run one second in the simulation, with varying numbers of discrete rod sections $n$. Simulation time steps of $\SI{0.0015}{s}$ is used as it balances accuracy and stability with computational time.

The increase in computational time for \verb!adapted! is small ($-1\% \text{ to } 3\%$) as compared to \verb!direct! ($2\% \text{ to } 15\%$). This difference is attributed to \verb!adapted! bypassing the $n+1$ distinct queries for the Jacobian at each simulation step. It is worth noting that the computational time increase for \verb!native! ($-3\% \text{ to } 2\%$) is smaller than \verb!adapted! as the latter goes through a more complex stiffness computation. A negative percent for some results could be attributed to the \verb!plain! model going through a more computationally expensive time step due to its less structure movement.

\subsection{Validation}
To guarantee accuracy of \verb!adapted! and \verb!native!, this section evaluates each model against available analytical solutions for elastic rods in the localized helical buckling~\cite{van2000helical} and Michell's instability tests~\cite{michell1889stability}. \verb!direct! is excluded from further evaluations as it has been tested to exhibit identical behavior to \verb!adapted!. Videos of the simulated validation (including unstable simulation of the native cable model) are available at: \url{https://youtu.be/QY0B5IChghw}.

\begin{figure}[t]
	\centering
	\includegraphics[width=0.45\textwidth]{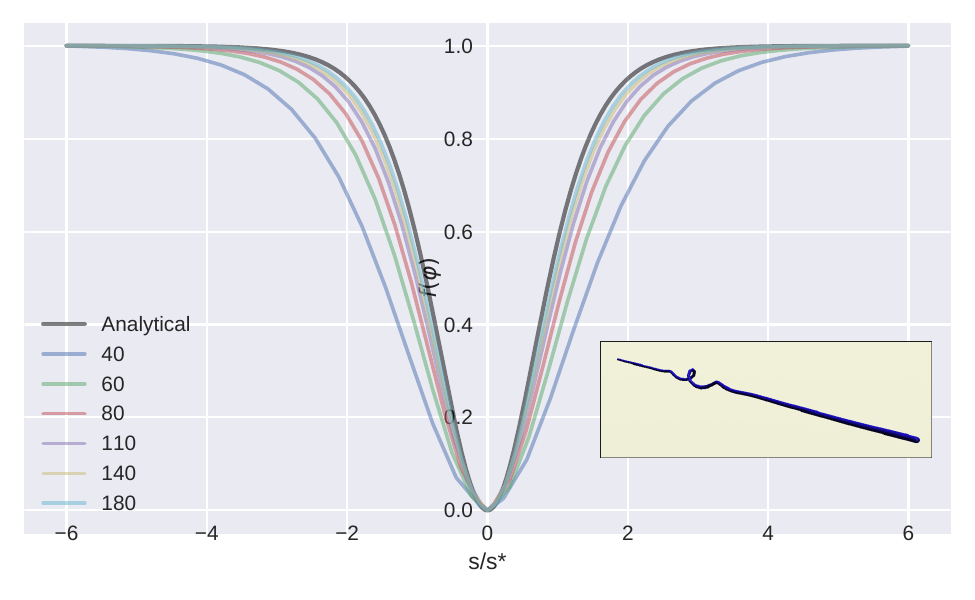}
	\vspace{-0.2cm}
	\caption{\textbf{Localized Helical Buckling Test} on a straight isotropic rod. A twist of 27 turns is imposed on one end of the rod. Properties of the rod are as follows: $L = 9.29$, $\alpha = 1.345$, $\beta = 0.789$. The ends are brought closer together by 0.3 units which causes helical buckling on the rod. Tests are carried out with different values of $n$ as shown in the legends. Results and render shown are of the \texttt{adapted} model. The plot converges to the analytical values as $n$ increases. }
	\label{fig:lhbtest}
	\vspace{-0.2cm}
\end{figure}

\paragraph{Localized Helical Buckling}
For a straight isotropic rod, localized helical buckling occurs when a twist is introduced to one rod end and the rods ends are brought towards each other, quasi-statically. The helix envelope is $f(\varphi) = \tanh^2(\frac{s}{s^*}) = \frac{cos(\varphi) - cos(\varphi_0)}{1 - cos(\varphi_0)}$, where $\varphi = cos^{-1}(\mathbf{t}.\mathbf{e}_x)$ is the angular deviation of the tangent away from the axis passing through the rod end points, $s$ is the rod arc length, and $s/s^* = (\frac{\beta m}{2\alpha}\sqrt{\frac{1 - cos(\varphi_0)}{1 + cos(\varphi_0)}}) s$. The maximal angular deviation of the rod from the tangent can be calculated as $\varphi_0 = max_s \varphi(s)$. $m$ is the overall twist over the rod length. A plot of the helix envelope, $f(\varphi)$ against the dimensionless arc length, $s/s^*$, is shown in Fig.~\ref{fig:lhbtest}. Results for both \verb!adapted! and \verb!native! are shown in Table~\ref{table:LHBresults}. We observe that \verb!adapted! converges towards the analytical solution with increasing $n$. The improved model accuracy of \verb!adapted! resulted in generally smaller average errors. It converged more consistently toward the analytical solution as compare to \verb!native! due to its improved stability, a factor that is attributed to its quasistatic treatment of the centerline twist and therefore lack of unnatural twist wave oscillations.

\begin{table}[h]
	\caption{Validation Results for Localized Helical Buckling Test}
	\vspace{-0.3cm}
	\label{table:LHBresults}
	\begin{center}
		\resizebox{\columnwidth}{!}{%
			\begin{tabular}{|>{\centering}m{25mm}|p{23mm}|p{12mm}|}
				\hline
				\multirow{2}{=}{Number of \newline Discrete Sections, $n$} & \multicolumn{2}{c|}{Average Error (2-norm error per data point)}\\
				\cline{2-3}
				& \texttt{native} & \texttt{adapted}\\
				\hline
				40 & 0.0389 & 0.0257\\
				\hline
				60 & 0.0624 & 0.0141\\
				\hline
				80 & 0.00692 & 0.00900\\
				\hline
				110 & 0.00673 & 0.00493\\
				\hline
				140 & 0.00397 & 0.00315\\
				\hline
				180 & 0.00348 & 0.00189\\
				\hline
			\end{tabular}
		}
	\end{center}
	\vspace{-0.4cm}
\end{table}

\begin{figure}[t]
	\centering
	\includegraphics[width=0.35\textwidth]{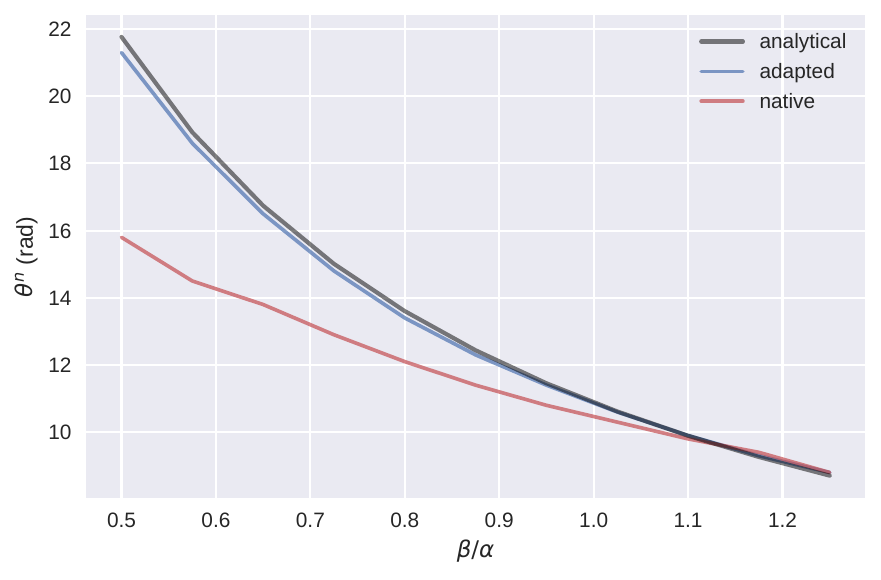}
	\includegraphics[width=0.12\textwidth, trim=0 -1cm 0 0]{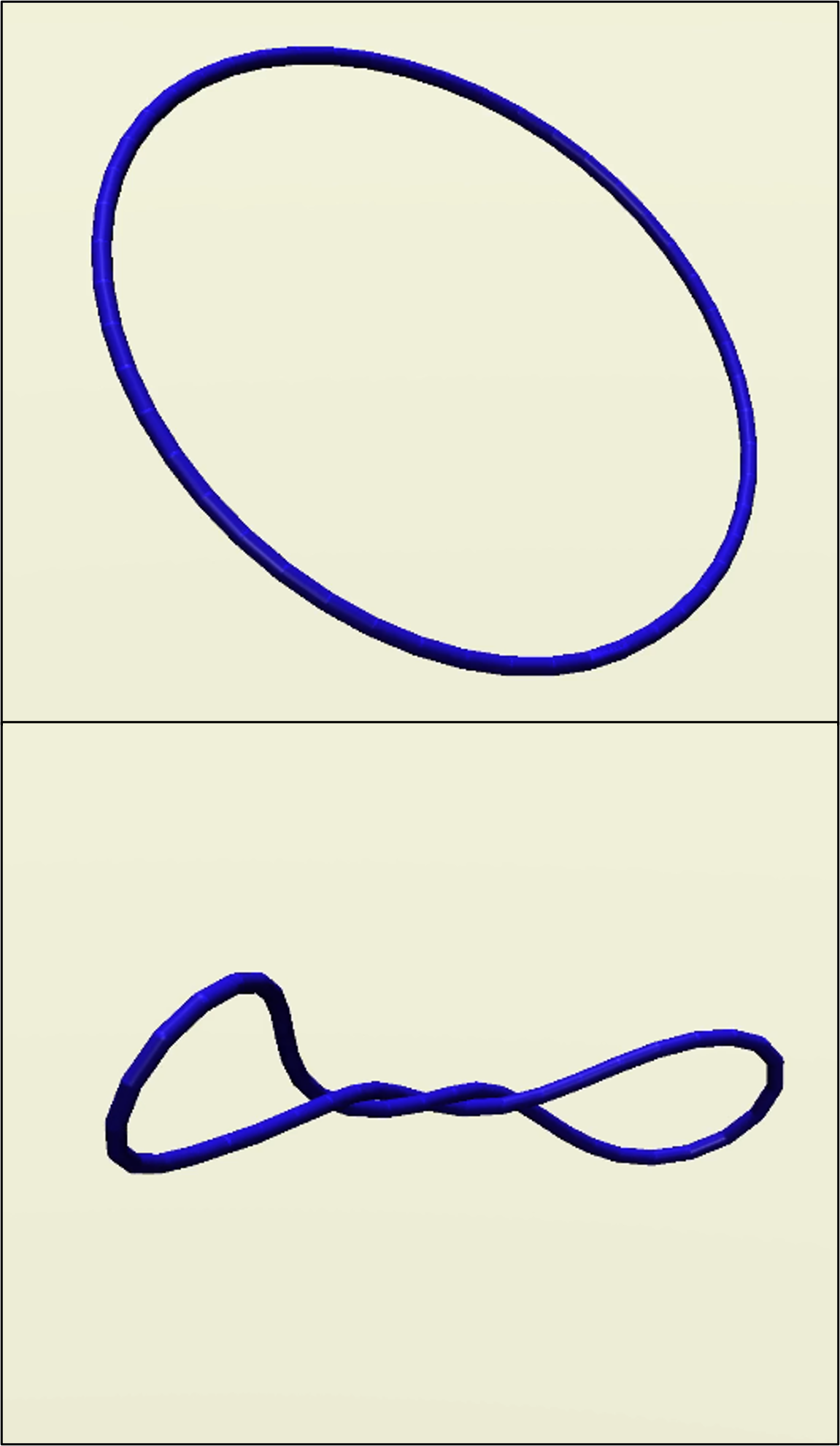}
	\vspace{-0.2cm}
	\caption{\textbf{Michell's Buckling Instability Test} on a straight isotropic rod with ends connected to form an elastic ring. One end is twisted a certain amount until a critical value is reached whereby the circular shape buckles into a non-planar shape in the presence of small external disturbances. $\alpha = 1$ and number of discrete rods, $n = 50$, while $\beta$ is varied to achieve different values of $\beta/\alpha$ for which the critical twist, $\theta^n$ depends. Renders shown are of the \texttt{adapted} model (top right: pre-buckling, bottom right: post-buckling).}
	\label{fig:mbitest}
	\vspace{-0.2cm}
\end{figure}

\paragraph{Michell's Buckling Instability}
By introducing and increasing twist along the tangent axis to an elastic rod loop, buckling occurs at a critical twist angle which can be determined analytically as $\theta^n_c = 2\pi\sqrt{3}/(\beta/\alpha)$ when the loop radius $R = 1$ \cite{goriely2006twisted}. This phenomenon shows the coupling between rod twist and equilibrium configuration. Fig.~\ref{fig:mbitest} shows the graph of critical buckling angle against $\beta/\alpha$ that was obtained for each simulation model. The average error defined as the 2-norm error per data point for \verb!adapted! of 0.0483 is almost an order of magnitude smaller than \verb!native! with 0.7725. \verb!adapted! provides results significantly closer to the analytical solution. Data of \verb!native! seems to exhibit a linear relation between $\beta/\alpha$ and $\theta^n_c$, which is inconsistent with a theoretical elastic rod. In Section~\ref{section:eval}, we will discuss in detail the possible shortcomings of \verb!native! that causes its failure and how \verb!adapted! accounts for these limitations.

\section{PARAMETER IDENTIFICATION}
\label{section:paramiden}
In this section, we introduce a novel parameter identification pipeline aimed at identifying the stiffness parameters of an elastic deformable linear object. The pipeline leverages two experiments -- the first will predict bending stiffness and the second will determine the $\beta/\alpha$ ratio. Parameter identification is accomplished through comparisons of the real experiments with a simulation environment which exactly replicates the real setup. We demonstrated our approach on three distinct DLOs -- white (silicone rubber), black (copper internals, polyvinyl chloride, PVC body), and red (copper internals, PVC body, nylon braided cover).

\subsection{Experimental setup}
\begin{figure}[t]
	\centering
	\includegraphics[width=0.30\textwidth]{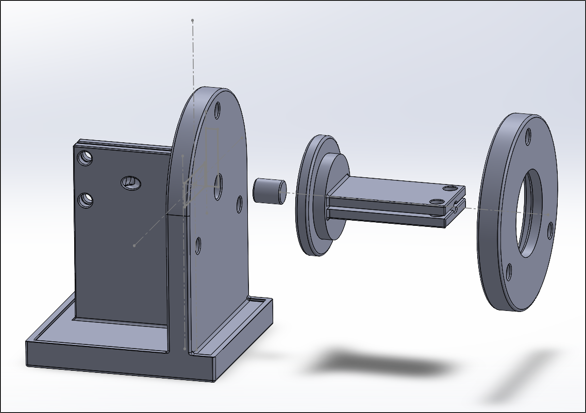}
	\hspace{-0.25cm}
	\includegraphics[width=0.166\textwidth]{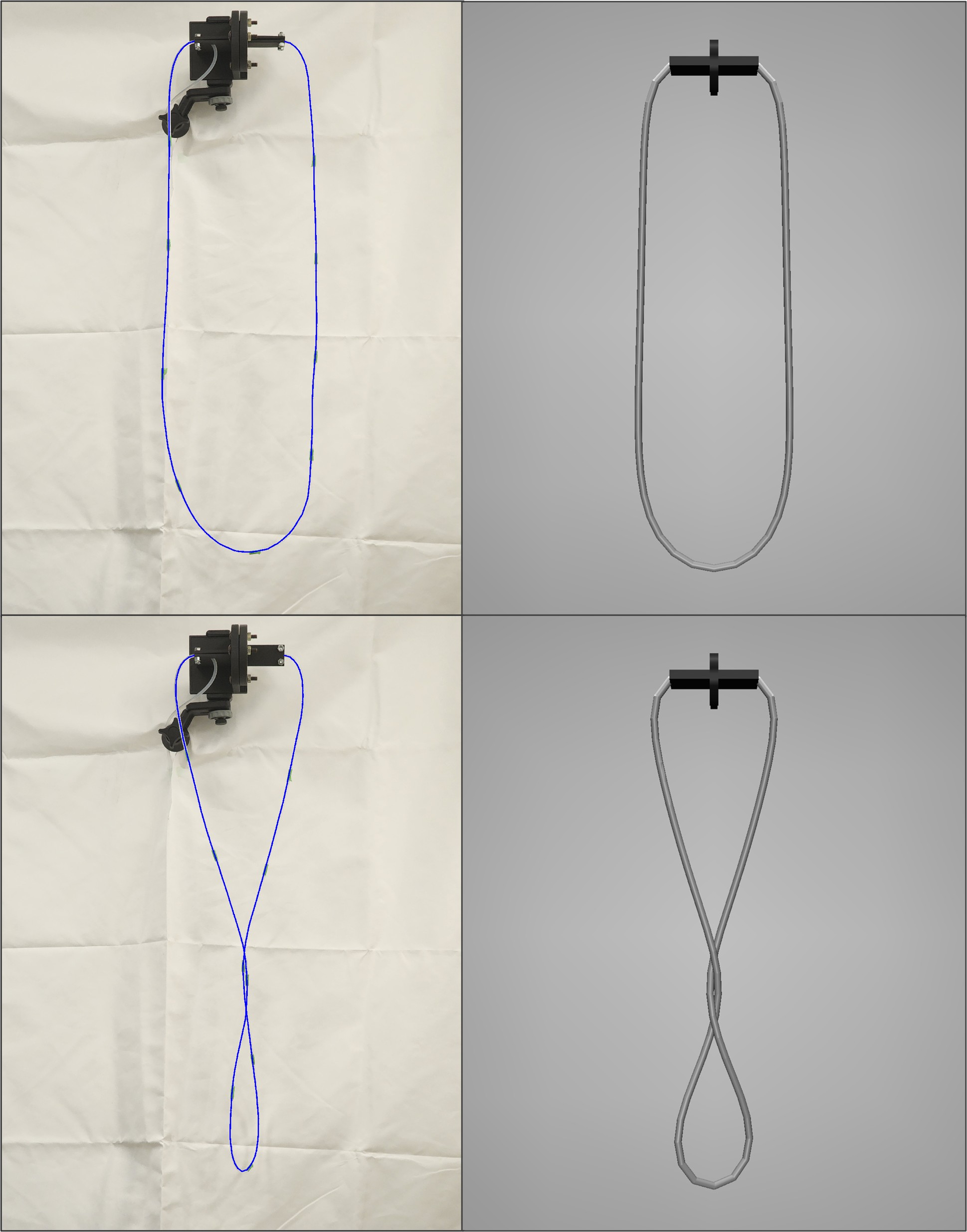}
	\vspace{-0.1cm}
	\caption{Design of the custom twist apparatus (CTA) used for parameter identification (left). All components can be 3-D printed and fastened with standard M5 nuts and bolts. The parameter identification setup for the white DLO, illustrating no twist (top right) versus critical twist (bottom right), are shown for both the real (highlighted in blue to improve visibility) and simulated environments. The CTA is being used to hold and introduce twist into the DLO.}
	\label{fig:piwhite}
	\vspace{-0.2cm}
\end{figure}

The experiments share a common setup featuring a set of 3-D printed apparatus and an Azure Kinect depth camera. Two ends of a $\SI{1.5}{m}$ DLO are fastened to the custom twist apparatus (CTA) which is affixed to a tripod using a phone-sized holder, and left to dangle in a loop. An exact replica of the setup is created in simulation. The CTA design is available on our Github repository for 3-D printing (STL format). To ensure effective DLO detection, green markers were placed at regular intervals along its length and a white sheet was placed behind the setup as shown in Fig.~\ref{fig:piwhite} along with the CTA design.

\subsection{Bending stiffness identification}
\subsubsection{Procedure}
The CTA is set to its neutral position ($0^\circ$ twist). The depth camera captures the DLO which is linearly spaced into $N=50$ nodes. DLO detection is manually executed. Following the general approach for parameter identification described in \cite{shutov2020sample}, the golden-section search algorithm compares real and simulated results to estimate the bending stiffness modulus $\alpha^*$ such that 
\begin{equation}
	\alpha^* = \arg\min_{\alpha} \left( \frac{1}{N} \sum_{i=1}^{N} \left\| \mathbf{x}_{\text{real}}(i) - \mathbf{x}_{\text{sim}}(\alpha, i) \right\|_2^2 \right)
\end{equation}
where $\mathbf{x}_{\text{real}}$ and $\mathbf{x}_{\text{sim}}$ are the real and simulated node positions, respectively, the latter having a dependence on $\alpha$.
\subsubsection{Evaluation}
The experiment was repeated $M=5$ times. Each time, the bottom end of the DLO was displaced $45^\circ$ towards the top right by $\SI{10}{cm}$, released, and allowed to arrive at its new equilibrium. To determine the suitability of our approach for parameter identification of each DLO, we calculated the normalized parameter cloud size \cite{shutov2020sample} across the experiments as follows,
\begin{equation}
	\label{eq:paramcloudsize}
	S = \frac{1}{M} \sum_{i=1}^{M} \frac{\lvert \bar{p} - p_i \rvert}{\bar{p}}\times100\%
\end{equation}
where parameter $p = \alpha_i$ is the bending stiffness moduli obtained from experiment $i$ and $\bar{p} = \bar{\alpha}$ is the average across $M$ experiments. The smaller the value of $S$, the more suitable the experiment is for parameter identification on the specific DLO sample, based on the consistency of the results. From the computed $S_{bend}$ shown in Table~\ref{table:paramiden}, we conclude that this parameter identification experiment is most suitable for use on the white DLO, likely because it exhibits the least plastic deformation.

\subsection{Twisting stiffness identification}
\subsubsection{Procedure}
The real critical twist angle $\theta_{c,real}$ is defined as the angle at which buckling first occurs (first instance of self-collision after loop folds into itself). To incite buckling in the dangling loop, twist is introduced to the CTA in increments of $5^\circ$ where one end is kept fixed while the other is axially rotated. Using a bi-section algorithm, we find the $\beta/\alpha$ ratio which gives $\theta_{c,sim} = \theta_{c,real}$ in the simulation, where $\alpha = \bar{\alpha}$ from the first experiment.
\subsubsection{Evaluation}
The experiment was repeated 5 times. The twist angle was returned to $0^\circ$ before each repeat. Using Eq.~(\ref{eq:paramcloudsize}) where $p = (\beta/\alpha)_i$, we calculated $S_{twist}$ for the twisting experiment. Results are shown in Table~\ref{table:paramiden}. We find that this experiment shares the same order of DLO suitability as the first experiment, with the white being the most suitable and having the most consistent results.

\begin{table}[h]
	\caption{Parameter cloud size for each experiment}
	\vspace{-0.2cm}
	\label{table:paramiden}
	\begin{center}
			\resizebox{0.40\textwidth}{!}{%
					\begin{tabular}{|c|c|c|c|}
							\hline
							\multirow{2}*{Experiment Type} & \multicolumn{3}{c|}{Parameter Cloud Size (\%)}\\
							\cline{2-4} &
							white & black & red\\
							\hline
							Bending & 2.182 & 6.064 & 7.553\\
							\hline
							Twisting & 0.155 & 1.625 & 2.297\\
							\hline
						\end{tabular}
				}
		\end{center}
	\vspace{-0.3cm}
\end{table}

\subsection{Evaluation of proposed pipeline}
Our approach is cheaper to deploy than current engineering approaches~\cite{bartholdt2021parameter,liu2023robotic}, requiring minimal equipment and sensors. It is also simpler to implement than learning approaches \cite{caldwell2014optimal} which require collection of a large amount of real data. This approach is more suited for parameter identification of rods which exhibit more elastic behaviors. The tested samples exhibited increasingly plastic deformations in the following order: white, black, and then red. Therefore, it is reasonable that suitability evaluations using the parameter cloud sizes $S_{bend}$ and $S_{twist}$ does place the DLOs in a reversed order when considering increasing suitability levels -- red, black, and then white. This pipeline is not meant to replace learning techniques for parameter identification. Instead, it can provide a warm start to these techniques, thereby reducing the amount of real training data required. After testing for consistency using the parameter cloud size, we next investigate the validity of the identified stiffness values using real experiments.

\section{REAL EXPERIMENTS}
\label{section:realexp}
To evaluate the performance of the novel parameter identification pipeline, real experiments were conducted where one end of a $\SI{0.40}{m}$ DLO was kept fixed while the other was held at 4 different poses by a Denso VS-060 robot arm. At each pose, the axial rotation at the fixed end held by the CTA was adjusted to introduce more twist into the system. Using the three DLOs from the previous section, real and simulated shapes were compared. The DLOs were discretized into 10 sections for comparisons. In the real experiments, an Azure Kinect was used for position detection along with linearly spaced green markers on the DLOs. Snapshots of the real and simulated experiments for the black DLO (visually more distinct than the white) are shown in Fig.~\ref{fig:realexpsnapshots}.

\begin{figure}[htp]
	\centering
	\setlength{\fboxsep}{0pt}
	\vspace{-0.2cm}
	\begin{tikzpicture}
	  	\node[inner sep=0pt] (bars) at (0,5) {\includegraphics[width=0.45\textwidth]{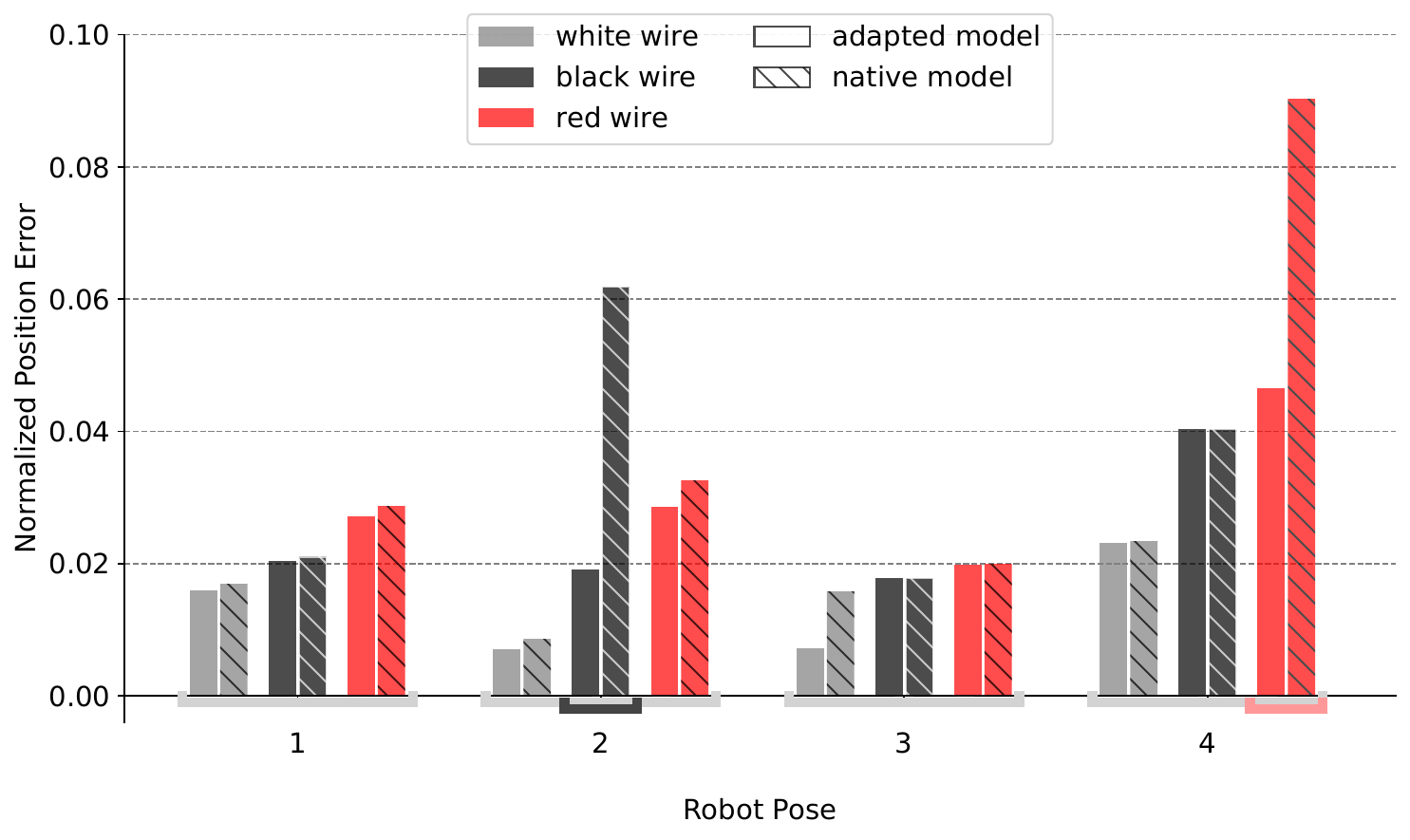}};
	    \node[inner sep=0pt] (img0) at (-2.89,1.2) {\fbox{\includegraphics[trim=17cm 6cm 50cm 30cm,clip,width=0.1605\textwidth]{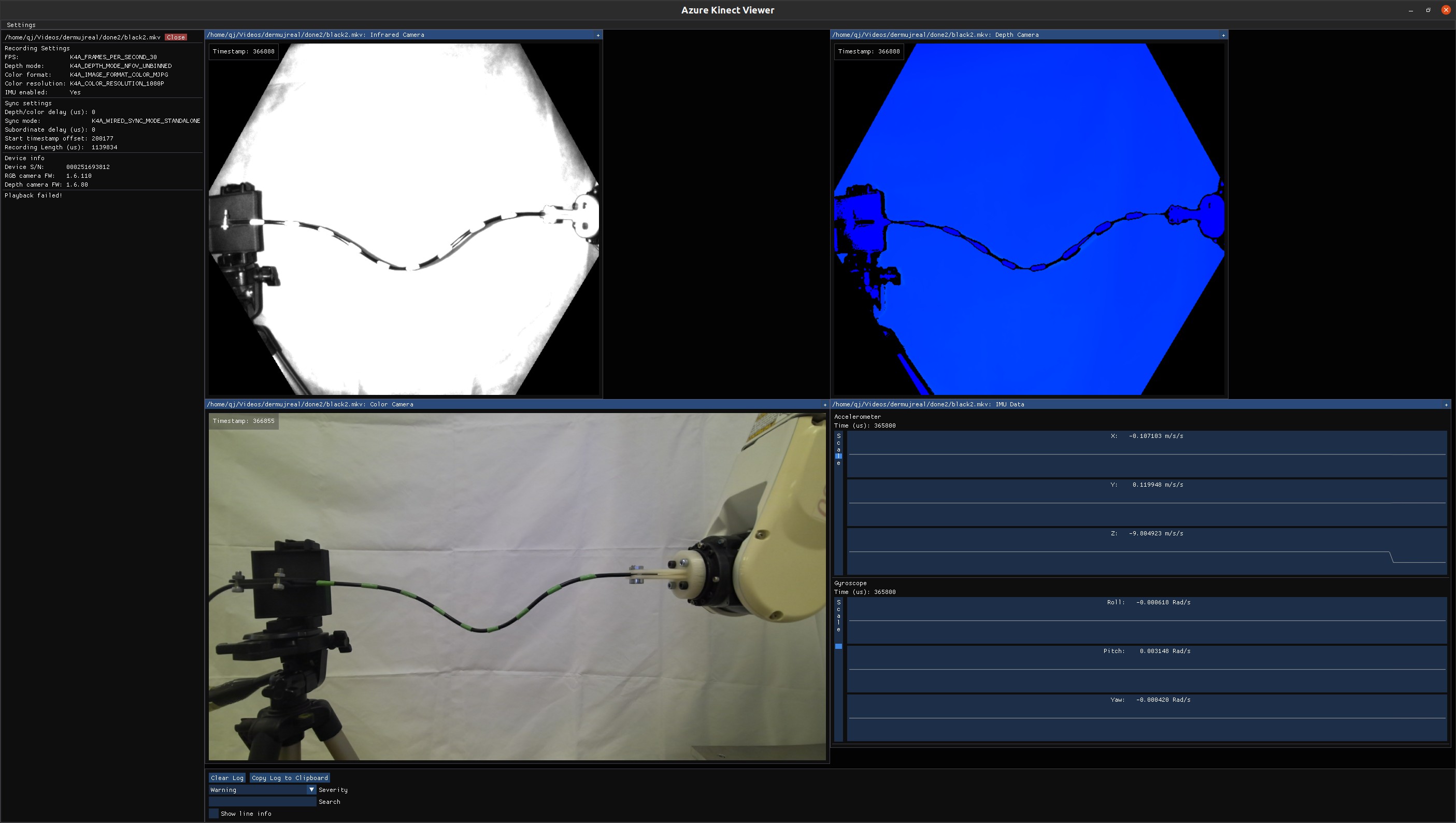}}};
	    \node[inner sep=0pt] (img1) at (0.0,1.2) {\fbox{\includegraphics[trim=20cm 9cm 20cm 9cm,clip,width=0.163\textwidth]{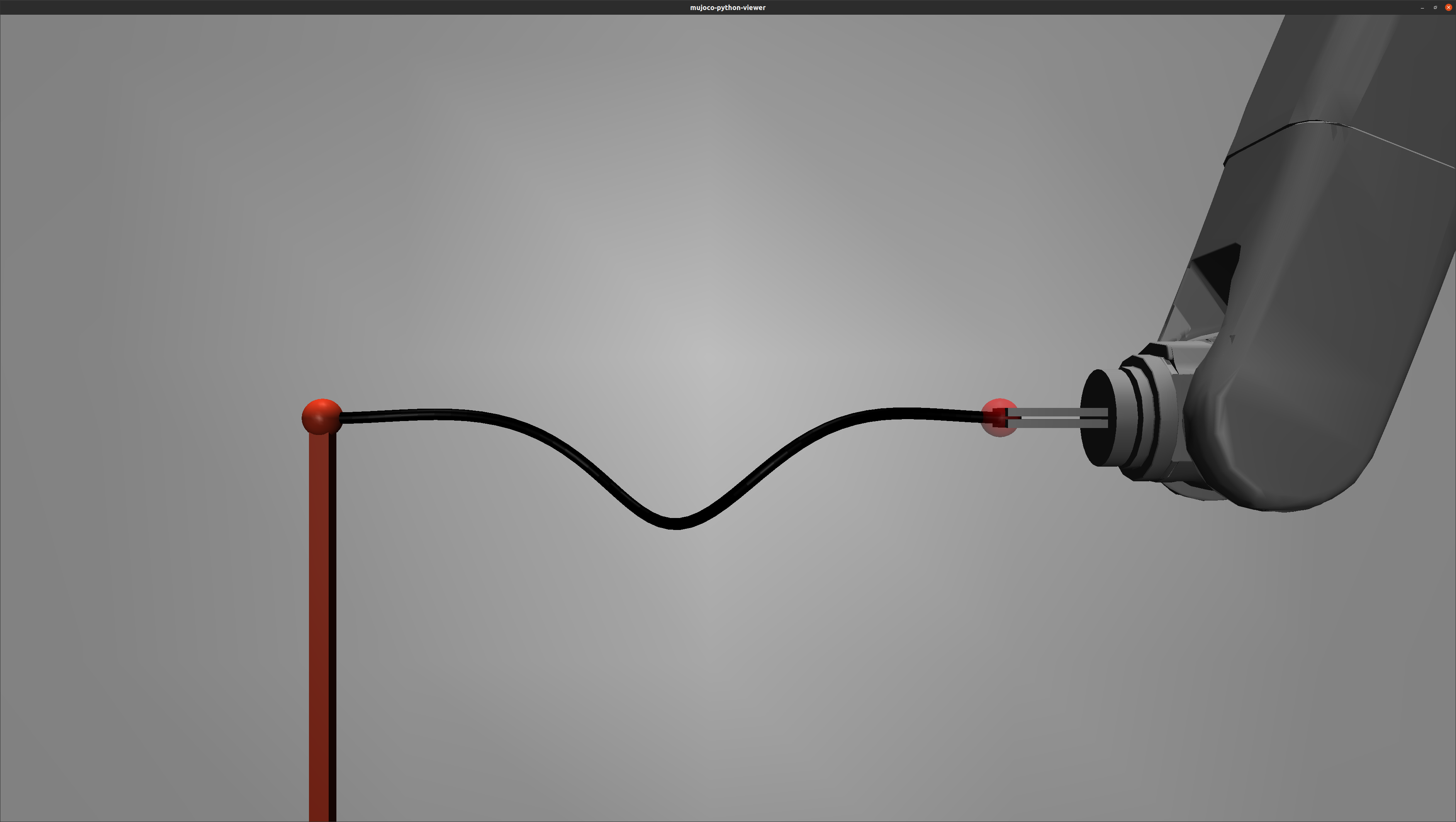}}};
	    \node[inner sep=0pt] (img2) at (2.92,1.2) {\fbox{\includegraphics[trim=20cm 9cm 20cm 9cm,clip,width=0.164\textwidth]{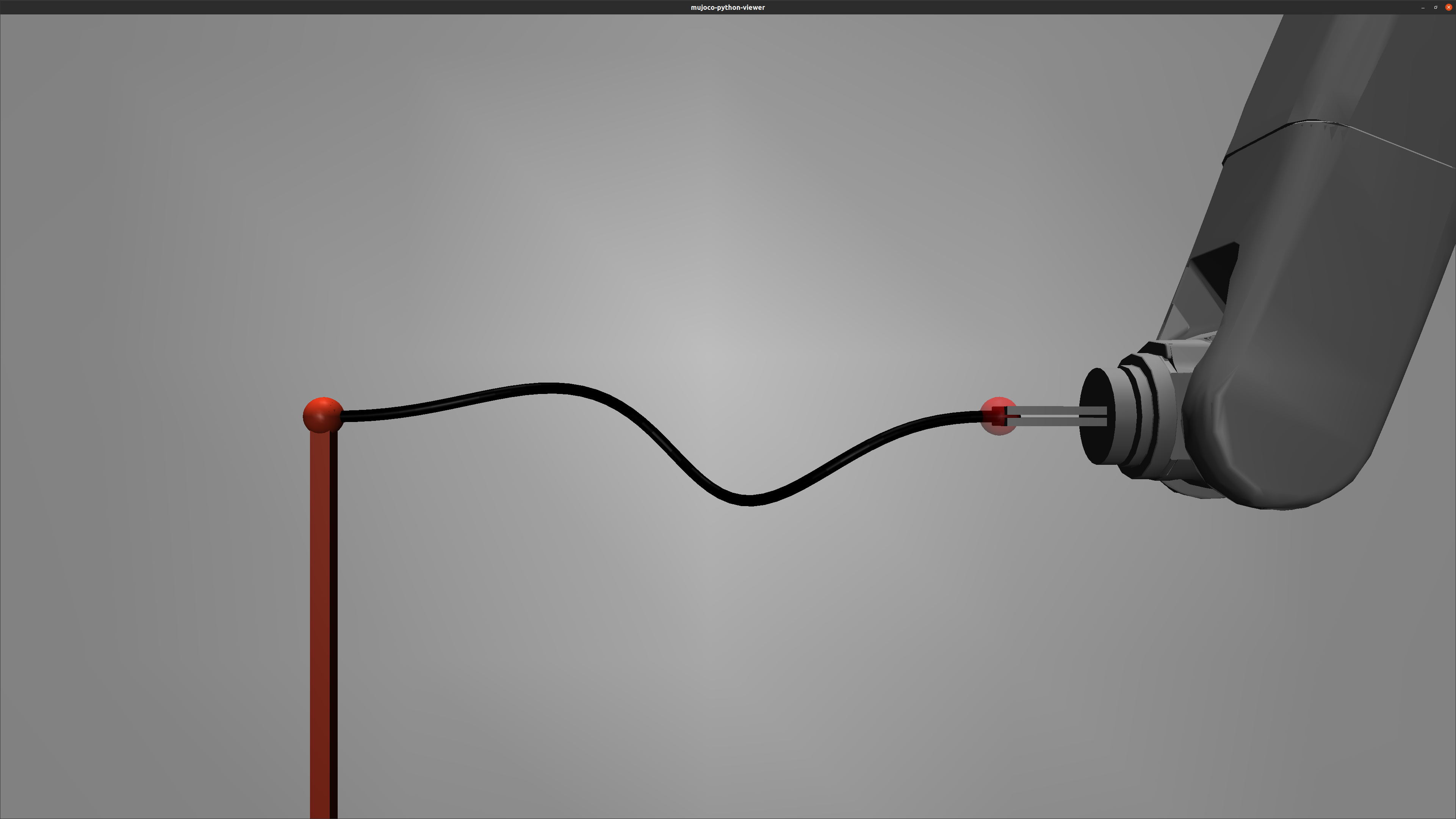}}};
	    \node[inner sep=0pt] (img3) at (-2.89,-1.2) {\fbox{\includegraphics[trim=15.7cm 5cm 48cm 29cm,clip,width=0.1605\textwidth]{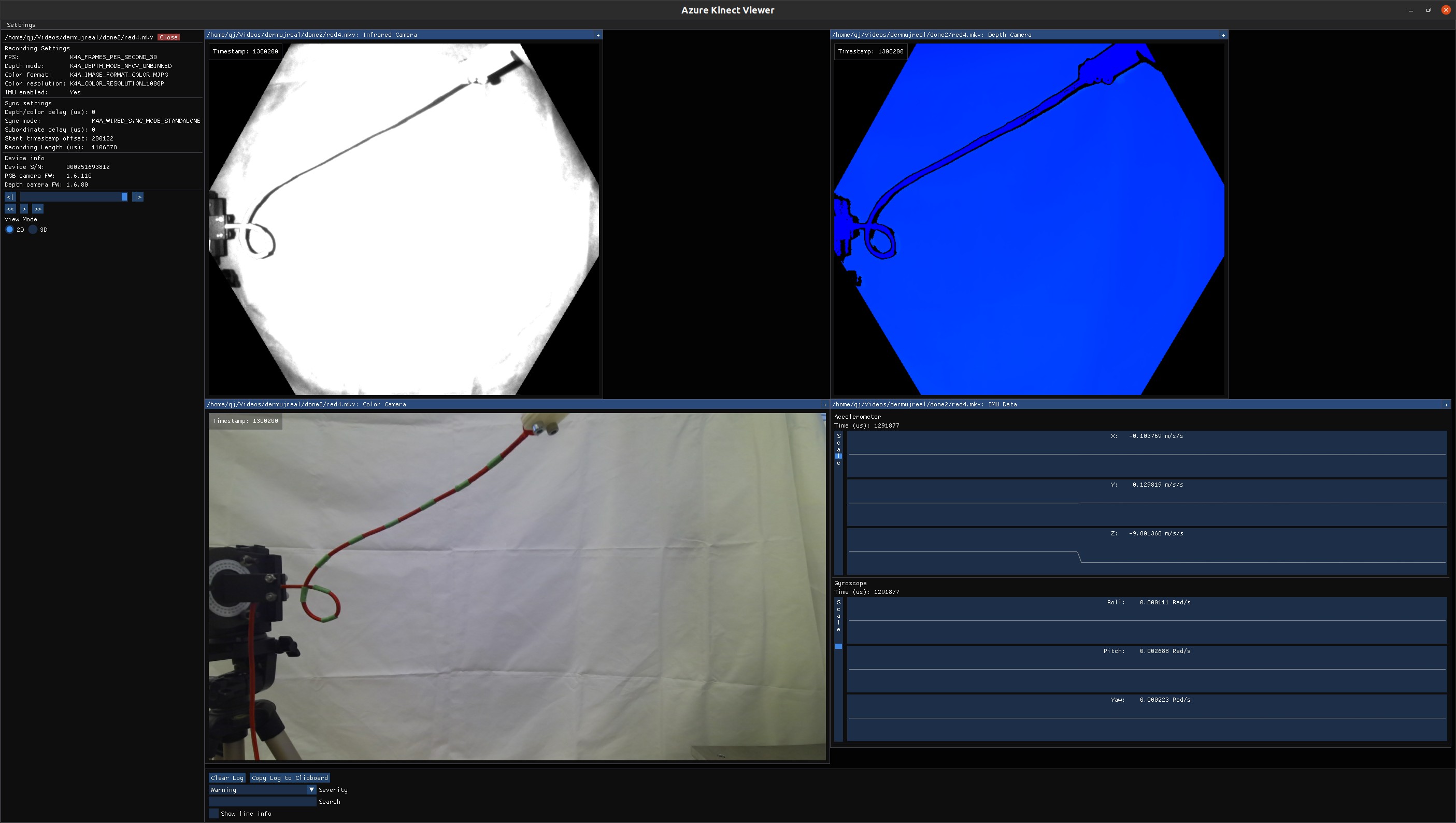}}};
	    \node[inner sep=0pt] (img4) at (0.0,-1.2) {\fbox{\includegraphics[trim=20cm 9cm 20cm 9cm,clip,width=0.163\textwidth]{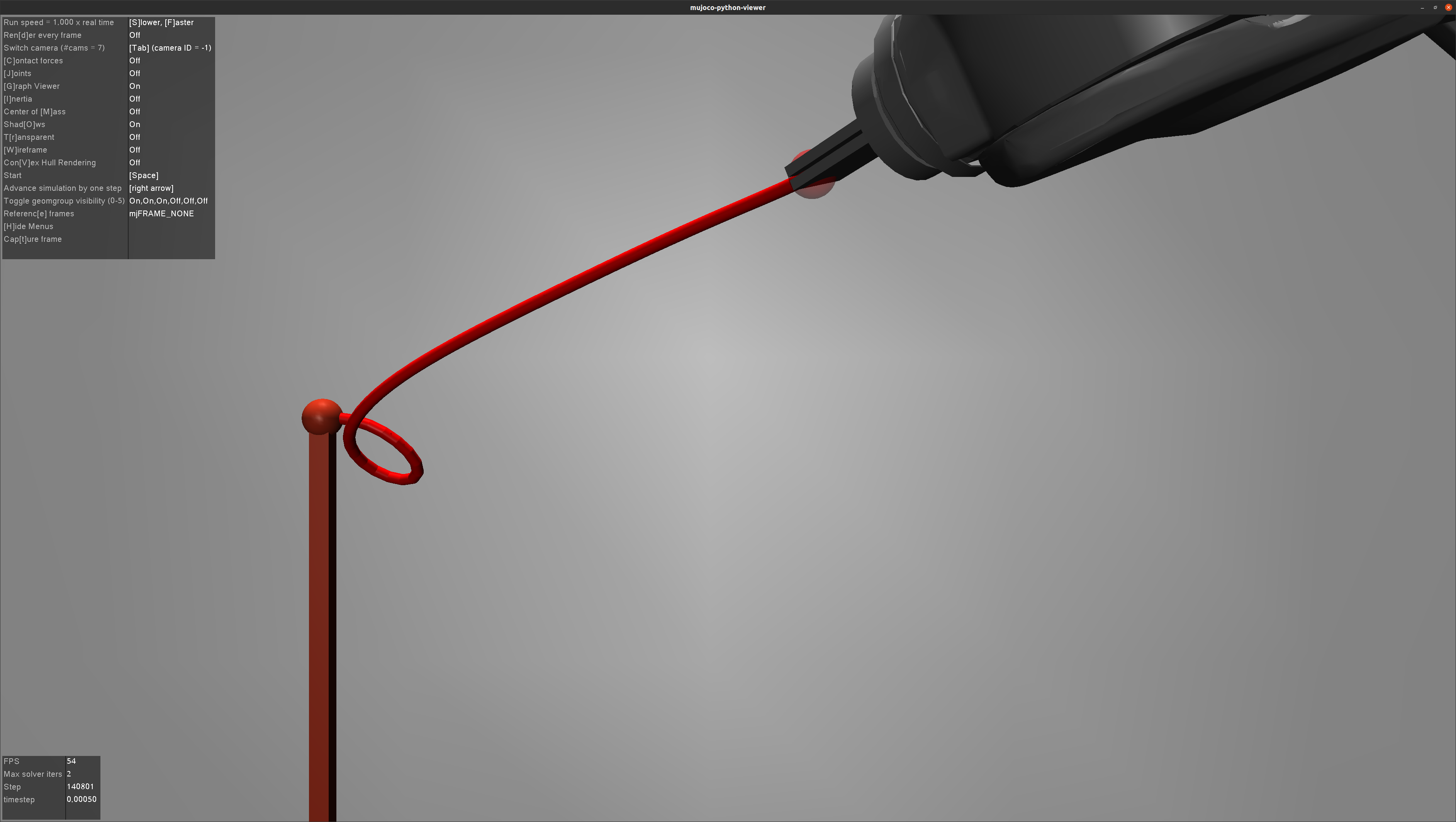}}};
	    \node[inner sep=0pt] (img5) at (2.92,-1.2) {\fbox{\includegraphics[trim=20cm 9cm 20cm 9cm,clip,width=0.164\textwidth]{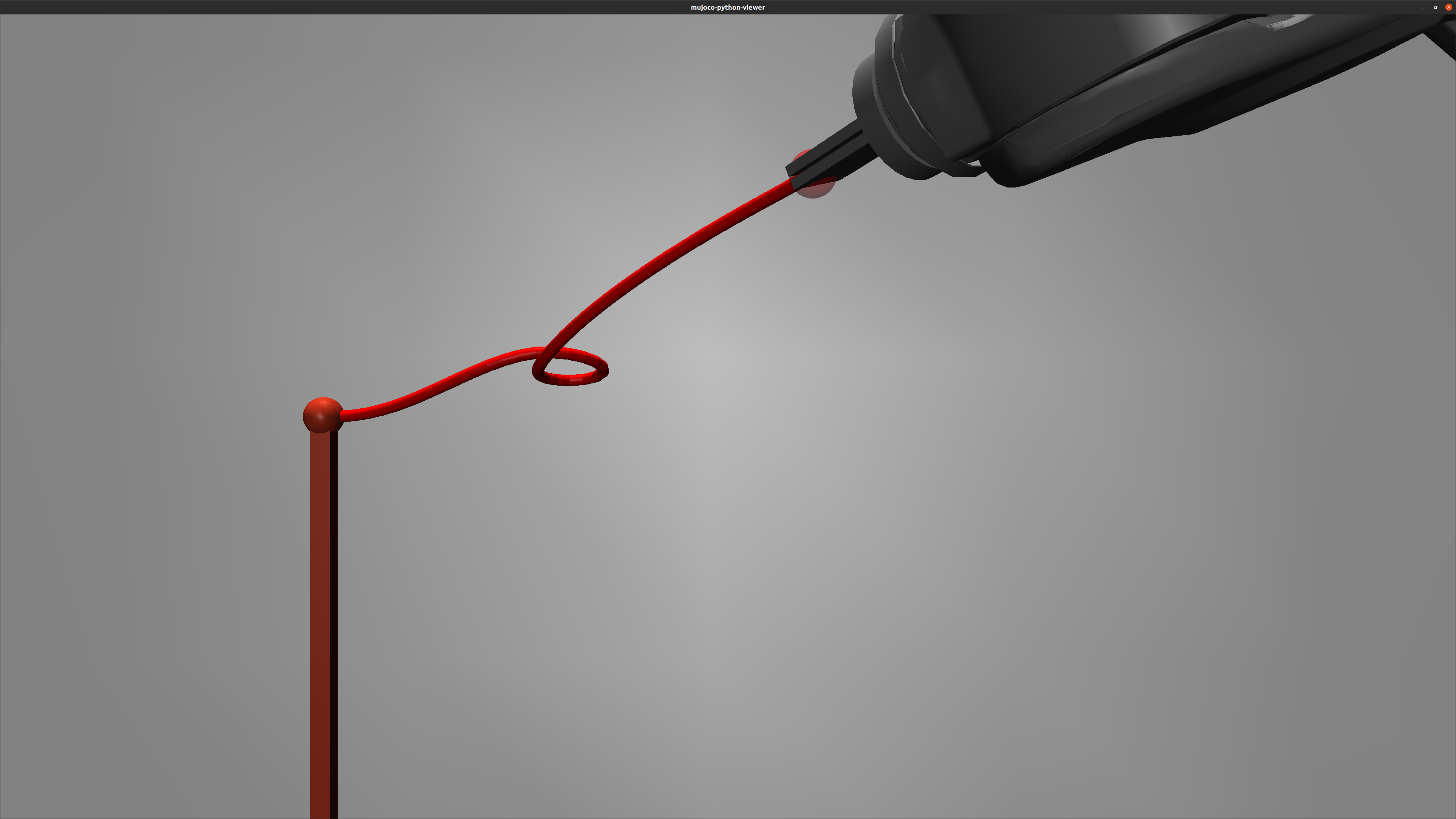}}};
	    
	    \node[anchor=south] at (bars.north) {\textbf{A}};
	    \node[anchor=south] at (img1.north) {\textbf{B}};
	    
	    \node[fill=black, fill opacity=0.5, text opacity=1, text=white, minimum width=0.3, rounded corners, inner sep=3pt, align=center]
	    (box) at (-2.89,0) {\small\verb!real!};
	    \node[fill=black, fill opacity=0.5, text opacity=1, text=white, minimum width=0.3, rounded corners, inner sep=3pt, align=center]
	    (box) at (0,0) {\small\verb!adapted!};
	    \node[fill=black, fill opacity=0.5, text opacity=1, text=white, minimum width=0.3, rounded corners, inner sep=3pt, align=center]
	    (box) at (2.92,0) {\small\verb!native!};

	    \coordinate (target) at (4.2,-0.19); 
		\coordinate (target2) at (3.9,2.17);
	    \draw[thick, red] (3.5,3.35) -- (4.21,3.35);
	    \draw[->, thick, red] ([xshift=0cm,yshift=0cm]4.2,3.35) -- (target);
	    
	    \draw[thick, black] (-0.38,3.35) -- (-0.38,3);
	    \draw[thick, black] (-0.39,3) -- (3.91,3);
	    \draw[->, thick, black] ([xshift=0cm,yshift=0cm]3.9,3) -- (target2);

		\node[draw=black, opacity=0.7, thick, sharp corners, fit=(img0) (img1) (img2), inner sep=1pt] {};
		\node[draw=red, opacity=0.7, thick, sharp corners, fit=(img3) (img4) (img5), inner sep=1pt] {};
	\end{tikzpicture}
	\vspace{-0.5cm}
	\caption{\textbf{a}: Normalized position error between the real and simulated environments for 4 different poses. The adapted model generally performs better than the native one. \textbf{b}: Significant shape differences in the same real experiments for the black wire in pose 2 (top row -- deformation of the native model protrudes above the horizontal plane from the grasp points) and the red wire in pose 4 (bottom row -- loop formed in the native model occurs near the middle rather than the end of the wire). For these cases, the adapted model (center column) performed better than the native model (right column) in terms of shape comparison to real wires (left column).}
	\label{fig:simvreal}
	\vspace{-0.3cm}
\end{figure}

From Fig.~\ref{fig:simvreal}, we observe that the adapted model is generally better at shape prediction for all DLOs. The normalized position error, calculated as the average 2-norm euclidean error of 11 ordered points, between real and simulated results was the smallest for the white DLO across all poses. This is likely due to it exhibiting greater elasticity than the others. Interestingly, we observed that cases of spiked position error (black in pose 2 and red in pose 4 shown in Fig.~\ref{fig:simvreal}(b)) only occurred for the native model. These large inaccuracies arose because of the unnatural twist wave oscillations in the native simulation model, where twist waves repeatedly travel back and forth along the DLO length. Without additional training from real data, our parameter identification pipeline along with accurate simulation was able to consistently predict shapes with position errors per node of $<5\%$ of total length.

\section{DISCUSSION}
\label{section:eval}
From the results, it is evident that \verb!native! leads in terms of computational time, but this comes at the expense of model accuracy. \verb!adapted! shows promising results in both validation tests and real experiments, with minimal increase in computational time. This section will examine possible reasons for the performance of each model and how \verb!adapted! is able to make up for the limitations of \verb!native!.

\subsection{Limitations of the native model}
In the localized helical buckling test, oscillations in the native simulated model caused inconsistent results which do not converge upon the theoretical solution as well or as consistently as the results from the adapted model. For Michell's buckling instability test, the stiffness ratio seems to be linearly related to the critical twist angle, a result which does not align with the analytical solution. Comparisons of real experiments with the native model generally exhibited worse performance than when the adapted model is used. These problems may stem from inconsistencies between the representation of joint stiffness in the native model and real-world dynamics. Another reason could be the native model's dynamic treatment of the material frame twist leading to unnatural twist wave oscillations (shown in the attached video). To ensure simulation accuracy, small time steps are required to accommodate the fast twist waves through the DLO. This significantly increases computational time. Also, appropriate twist damping would have to be employed in the joints to ensure twist waves travel realistically. This poses a problem in MuJoCo as ball joints use only a single damping value for all axes of rotation. Tuning the damping parameter for twist is nontrivial due to its coupled effects with bending in the DLO.

\subsection{Advantages of the adapted model}
The generalized coordinates representation of DER theory removes unnecessary computational load, enhancing computational speed when compared to directly application of the theory. Because of the slender shape of a DLO, twist waves can be assumed to travel instantaneously. This quasistatic treatment of the centerline twist removes the need for axial rotational damping, mitigating the problem of coupled bend and twist damping in MuJoCo's ball joint. This mitigates the issue of unnatural twist wave oscillations and improves simulation stability.


\section{CONCLUSION}
To accurately simulate a DLO, we use the mathematical theory of DER integrated into a custom DLO model in MuJoCo. Our adapted model utilizes the properties of ball joints to efficiently derive a generalized coordinate representation of stiffness using force-lever analysis for use in MuJoCo. By adopting this model over the native cable model in MuJoCo, we avoid the phenomenon of unnatural twist wave oscillations and inaccurate stiffness representation which result in potentially severe inaccuracy of the simulation. Validation test results and comparisons with real experiments confirm our contribution. We present a easy-to-implement parameter identification pipeline and evaluate its effectiveness on three distinct real DLOs. We then compare the identified DLOs with their simulated counterparts for both the native and our adapted model, and conclude that our adapted model has a generally better shape prediction performance for the DLOs tested. Our work has potential contributions in reducing real training samples required for machine learning of dynamics and shape control task by providing a warm start estimate of the DLO stiffness parameters. 

\subsection*{Limitations and future work}
Our work would benefit from more rigorous testing on end-to-end robotic manipulation tasks. Future work could focus on demonstrating the model's effectiveness in improving control and planning for DLO manipulation. Further refinement of the parameter identification pipeline could involve integrating automated DLO position detection \cite{zhaole2023robust}, enabling a more finely discretized DLO for more detailed position comparisons. In addition, we intend to investigate the simulation of DLOs which experience plastic deformation and carry out parameter identification in a similarly straightforward manner.

\addtolength{\textheight}{-0cm}   


\end{document}